\documentclass{article}
\usepackage{arxiv}

\usepackage[english]{babel}
\usepackage[utf8]{inputenc} 
\usepackage[T1]{fontenc}    
\usepackage{url,hyperref}   
\usepackage{booktabs}       
\usepackage{amsfonts}       
\usepackage{nicefrac}       
\usepackage{microtype}      
\usepackage{lipsum}
\usepackage{graphicx,subfigure}
\usepackage[table,xcdraw]{xcolor}
\usepackage{tablefootnote}
\usepackage{listings}

\graphicspath{ {./images/} }

\hypersetup{
    colorlinks=true,
    linkcolor=blue,
    urlcolor=blue,
    citecolor=black,
}

\newcommand{\toolName}{{AskSport}}
\title{\toolName: Web Application for \\
Sports Question-Answering}

\author{
 {Enzo Baraldi de Onofre} \\
  Faculty of Computing\\
  Federal University of Uberlandia, Brazil\\
  \href{mailto:enzo.onofre@ufu.br}{enzo.onofre@ufu.br} \\
  \And
{Leonardo Mauro Pereira Moraes} \\
  Institute of Mathematics and Computer Sciences \\
  University of Sao Paulo, Brazil\\
  \href{mailto:leonardo.mauro@usp.br}{leonardo.mauro@usp.br} \\
  \And
 {Cristina Dutra de Aguiar} \\
  Institute of Mathematics and Computer Sciences \\
  University of Sao Paulo, Brazil\\
  \href{mailto:cdac@icmc.usp.br}{cdac@icmc.usp.br}
}

\begin{document}
\maketitle

\begin{abstract}
This paper introduces \toolName\footnote{\href{https://www.loom.com/share/e67861d2264040279e873a5f170d16ee?sid=e8640706-5625-4543-b52e-1e8c02a99eaa}{Demonstration of \toolName\ (pt-br).}}, a question-answering web application about sports. It allows users to ask questions using natural language and retrieve the three most relevant answers, including related information and documents. The paper describes the characteristics and functionalities of the application, including use cases demonstrating its ability to return names and numerical values. \toolName\ and its implementation are available for public access on HuggingFace\footnote{ \href{https://huggingface.co/spaces/leomaurodesenv/qasports-website}{Implementation of \toolName\ (huggingface.co).}}.

\end{abstract}

\section{introduction}
\label{sec:introduction}

Sport is a topic of great interest to society, as it encompasses cultural, social, and economic aspects. Different users, such as fans, coaches, and media professionals, are constantly looking for sports data for the most different purposes. For example, fans can obtain details on teams; coaches can evaluate the performance of players; and media professionals can analyze historical records in search of better players~\cite{jardim:2023:qasports-dataset}.

To meet this demand, there is a need for applications that access sports data and users query. The applications must also return coherent and structured answers in order to be understandable and useful, so that users can use them according to their needs. The main challenge is that these applications must incorporate question-answering algorithms~\cite{MISHRA2016345}.

These algorithms use advanced natural language processing (NLP) and machine learning techniques to understand and answer questions expressed in natural language. They analyze each question in order to understand the context in which it was asked, the subject being addressed, and its intention. In addition, the question-answering algorithms include a document retriever to query documents using similarity and a document reader to generate relevant answers using the retrieved documents. In this context, documents refer to large volumes of text that are used as a knowledge base.

On the web, question-answering applications can play a significant role as virtual systems for seeking knowledge on a variety of topics, such as sports. Within this context, this article presents \toolName, a question-answering web application about sports. The application enables users to ask questions using natural language and returns the three most relevant answers, along with related information. Its distinctive features are:
\begin{itemize}

\item Implementation of techniques used by question-answering algorithms for retrieving and reading documents using BigQA architecture as blueprint~\cite{moraes:2023:big-qa-architecture}.

\item Access to basketball sports data obtained from the QASports~\cite{jardim:2023:qasports-dataset} dataset, which represents the state-of-the-art currently available in the literature for sports questions and answers.

\item Providing an interactive interface that follows accessibility standards.

\end{itemize}

The article is organized as follows.
The~\ref{sec:related-works} section summarizes related work.
The~\ref{sec:methodology} section details the characteristics of \toolName.
The~\ref{sec:demonstration} section describes scenarios for using the application.
Section~\ref{sec:conclusion} concludes the article.

\section{Related Works}
\label{sec:related-works}

As this is a recent topic, there are few web applications based on question-answering algorithms. CHIM~\cite{schaffer:2022:let-ai-chatbots} allows users to ask questions about works of art in a museum. Unlike \toolName, which uses a document retriever and reader engine, CHIM performs similarity queries on a manually generated knowledge base, in a similar way to database queries. NLP techniques are only used to improve the relevance of the answers.


FALQU~\cite{mansouri:2023:falqu} uses the Law Stack Exchange lawyer question platform. It converts questions into title and content format. Then, given a document and a question, it finds the answer snippets most relevant to the question. FALQU uses only the document reader, while \toolName\ uses both document retriever and reader algorithms, providing more complete functionality. Another difference between FALQU and \toolName\ is the domain, namely legal and sports.

For completeness sake, the purposes of \toolName\, traditional search engines like Google and the ChatGPT model are different. \toolName\ is aimed exclusively at the sports domain, incorporating inherent knowledge and language, as well as using a sports dataset. These features aim to meet the needs of the application within the domain in question.

\section{\toolName}
\label{sec:methodology}

\subsection{Implementation Pipeline}
\label{subsec:arquitetura}

The core components of \toolName\ are the document retriever and reader, which represent techniques employed by question-answering algorithms. Another component is the QASports~\cite{jardim:2023:qasports-dataset} sports dataset. It contains the following collections of data on soccer, American football, and basketball: (i) web pages cleaned and stored in JSON files; (ii) context files stored in CSV format; and (iii) questions and answers stored in the context-question-answer triples format. 

\begin{figure}[t!]
\centering
\includegraphics[width=0.9\textwidth]{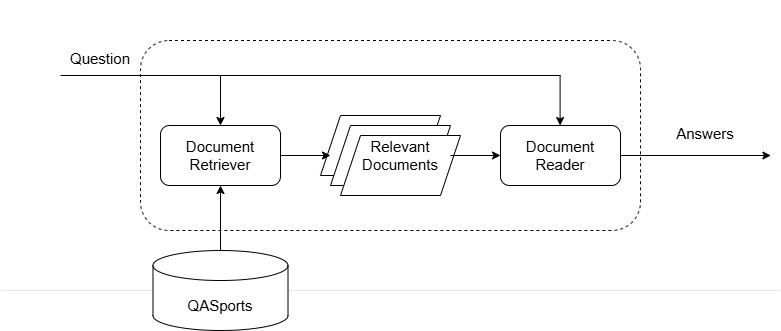}
\caption{Implementation Pipeline of \toolName.}
\label{fig:arquitetura}
\end{figure}

Figure~\ref{fig:arquitetura} shows the implementation pipeline of \toolName. Given a \textit{question} written in natural language, the \textit{document retriever} accesses the QASports dataset to perform the search. This search aims to find useful information in response to the question asked. It identifies and retrieves the ten \textit{most relevant documents}. Subsequently, the \textit{document reader} examines the retrieved documents, analyzing the information obtained using advanced natural language processing techniques to understand the context and semantics of the text. Finally, a result is produced to be returned.

\toolName\ returns the three most relevant \textit{answers} to the question. For each answer, the following contextual information is also returned: (i) confidence metric in the provided answer; (ii) title of the document used to generate the answer; and (iii) \textit{link} to the web page of the document.

\subsection{Interface and Capabilities}
\label{subsec:interface}

\begin{figure}[t!]
    \centering
    \includegraphics[width=0.9\textwidth]{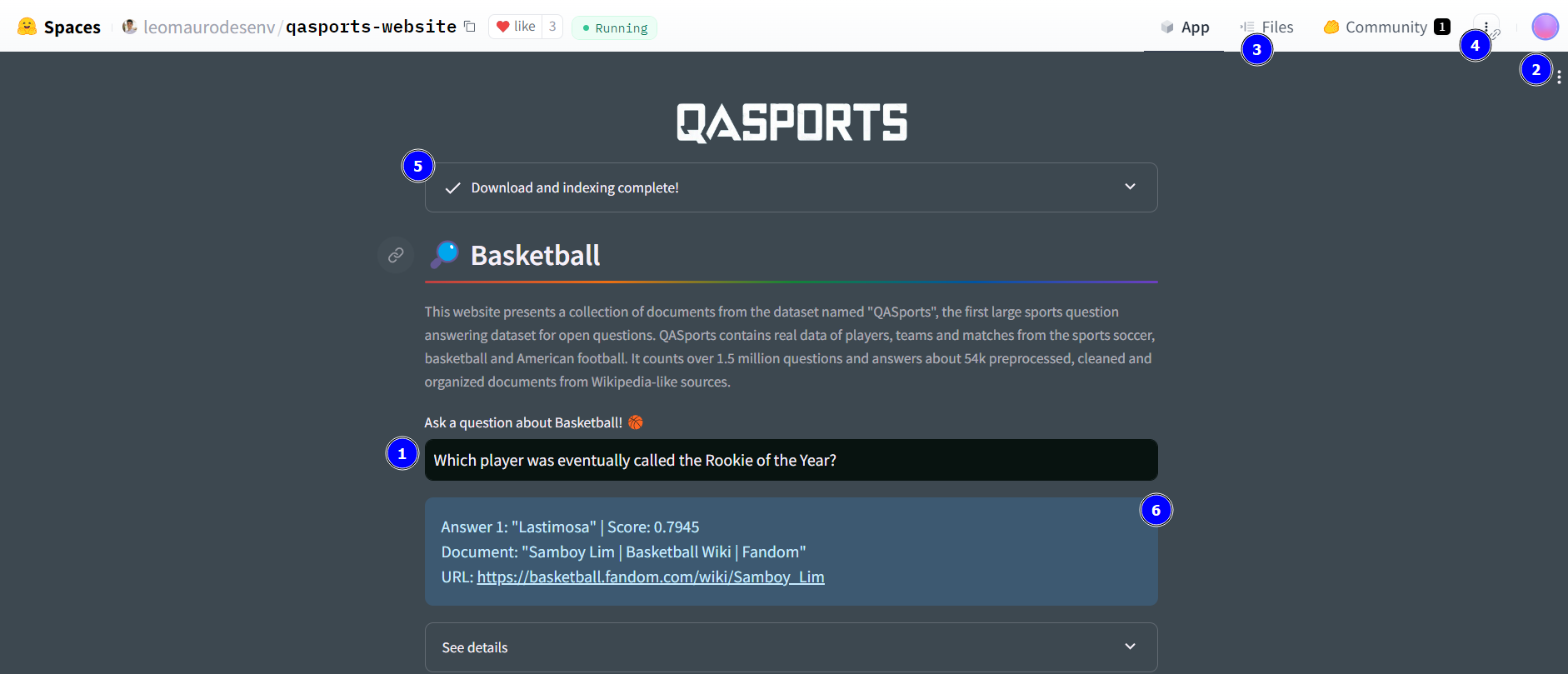}
    \caption{\toolName\ Interface. Demonstrating key capabilities.}
    \label{fig:website2}
\end{figure}

\begin{figure}[t!]
\centering
\subfigure[Aba 2: settings]{
\includegraphics[width=4cm]{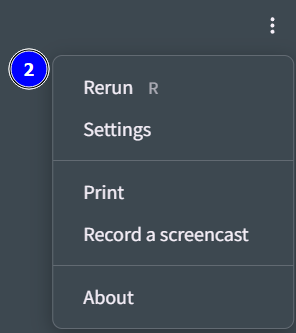}
\label{fig:subfigaba2}
}
\subfigure[Aba 4: QASports]{
\includegraphics[scale=0.5]{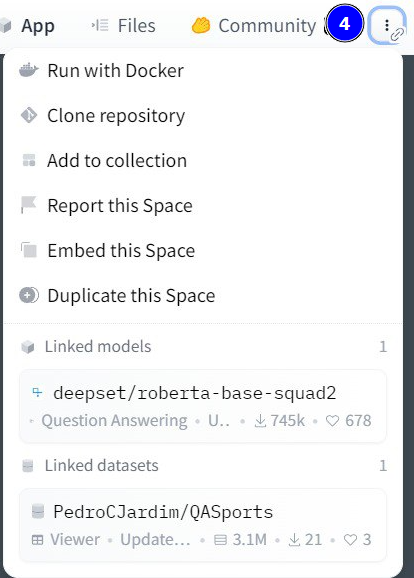}
\label{fig:subfigaba4}
}
\label{fig:abasOpcoes}
\caption{
Details of Tabs 2 and 4 of \toolName.}
\end{figure}

The \toolName\ interface (Figure~\ref{fig:website2}) was designed to offer an interactive and informative experience for users. Its functionalities are:
\begin{itemize}
\item \textbf{1:} Box in which the user can type the question using natural language.
\item \textbf{2:} Tab to manipulate the general settings of the interface (Figure~\ref{fig:subfigaba2}).
\item \textbf{3:} Tab to access the implementation code of \toolName.
\item \textbf{4:} Tab to access the QASports dataset (Figure~\ref{fig:subfigaba4}).
\item \textbf{5:} Box with the application status.
\item \textbf{6:} Box describing the three most relevant answers and their related contextual information (\textit{score}, \textit{document}, and \textit{URL}, as described in Section~\ref{subsec:arquitetura}). This information is displayed when accessing the ``\textit{See details}'' tab. The answers are ordered in descending order by the value of the metric. In Figure~\ref{fig:website2}, only one answer is displayed due to page limitations. When the user writes a text that does not make sense and, therefore, no answers are returned for the question, this box shows the phrase ``We do not have an answer for your question''.
\end{itemize}

To meet the needs of users with color blindness, the default colors of the interface comply with the \textit{Web Content Accessibility Guidelines 2.0}\footnote{\href{https://www.w3.org/TR/WCAG20/}{https://www.w3.org/TR/WCAG20/}}. The colors used are: (i) \textit{Light Gray} as the primary color (\#afbac2); (ii) \textit{Dark Slate Blue} as the background color (\#3d4850); (iii) absolute black as the secondary background color (\#081310); and a shade of white, called \textit{Ghost White}, as the text color (\#f5eff8). The color codes are in hexadecimal RGB format. In addition, the interface also has two other traditional themes to cater to different types of users: light and dark. This functionality can be found on Tab 2 (Figure~\ref{fig:subfigaba2}).

\subsection{Question-Answering}
\label{subsec:implementacao}

BM25~\cite{robertson:1976:bm25-algorithm} was used as \textit{document retriever}. It is a retrieval model that calculates the relevance of documents based on the frequency of terms. BM25 was chosen because it has provided better performance for retrieving data from QASports when compared to other models~\cite{moraes:2023:big-qa-architecture}. The \textit{document reader} was implemented using RoBERTa~\cite{liu:2019:roberta-model}, a natural language model that extends the BERT model. RoBERTa was selected for its efficiency in extracting relevant answers to general domain questions. In the current implementation, basketball data available in QASports is used.

The algorithms for document retriever and reader were implemented using Haystack\footnote{\url{https://haystack.deepset.ai/}}, a library for question-answering in the Python programming language. Designed to be modular and scalable, it allows developers to incorporate sophisticated search functionalities into their applications.
Regarding the implementation of the interface (section~\ref{subsec:interface}), Python version 3.10 was chosen due to its popularity and wide range of available libraries. The construction of the interactive \textit{front-end} of the interface was done using Streamlit\footnote{\url{https://streamlit.io/}}, since it facilitates the creation of websites.

We employed HuggingFace\footnote{\url{https://huggingface.co/}} as free hosting platform. It offers comprehensive and efficient solutions for developing and implementing NLP applications. For example, HuggingFace Models provides access to a vast library of pre-trained models, allowing developers to quickly integrate advanced question-answering capabilities into their applications. While HuggingFace Spaces enables simple and scalable hosting, allowing the publication of NLP applications.


\section{Applications for \toolName}
\label{sec:demonstration}

This section outlines various use case scenarios, emphasizing that the insights gained are more qualitative than quantitative. The scenarios are detailed below:

\noindent \textbf{Scenario 1.} The user wants to receive as an answer the name of a player who received a certain award. Therefore, they write the following question: ``which player was eventually called the Rookie of the Year?''

\noindent \textbf{Scenario 2.} The user wants an answer about a number of titles a certain team has. Therefore, they write the following question using natural language: ``How many titles do the NBA Warriors have?''


\noindent \textbf{Scenario 3.} The user wants to investigate a more comprehensive aspect of the basketball sports scene. Therefore, they write the following question: ``Who is considered the best basketball player in history?''

The answers obtained for each scenario and their respective metrics (score) are shown in Table~\ref{tab:cenarios}. Users can use these answers for the most different purposes, according to their needs.
The quality of the answers depends on the quality of the models used in the question-answering pipeline. In fact, in question-answering applications the users should analyze the corresponding documents to form their own opinions. For example, the question in Scenario 3 does not delimit a year or sport category, returning varied answers that should be analyzed. In this sense, \toolName\ not only returns each answer and its associated metric, but also the associated documents.  

\begin{table}[t!]
\centering
\begin{tabular}{lccr}
\hline
\textbf{Case} & \textbf{Question} & \textbf{Answer} & \textbf{Score}\\ \hline\hline
1   & Which player was & Lastimosa &  0.7945 \\ 
    & eventually called the & James & 0.7198 \\ 
    & Rookie of the Year? & Glenn Robinson & 0.6899 \\ 
\hline
2   & How many titles do the & seven &   0.7897 \\ 
    & NBA Warriors have? & three times & 0.7677 \\ 
    &  & two & 0.6377 \\ 
\hline
3   & Who is considered & Wilton Norman Chamberlain &  0.7978 \\ 
    & the best basketball & Aaron Jamal Crawford & 0.7108 \\ 
    & player in the history? & Earvin Magic Johnson & 0.6684 \\ 
\hline \hline
\end{tabular}
\caption{Questions and answers for different usage scenarios of \toolName.}
\label{tab:cenarios}
\end{table}

\section{Conclusion}
\label{sec:conclusion}

This article presented the \toolName\ application, which enables users to ask questions about sports using natural language and receive the three most relevant answers, including related information such as the documents in which the answers can be found. Its implementation, available on HuggingFace, was executed in Python and used BM25 as a document retriever and RoBERTa as a document reader, in addition to the QASports sports dataset. Additionally, the interactive interface follows accessibility standards.
Due to its implementation pipeline, \toolName\ is flexible in the sense that other algorithms can be employed as a document retriever and reader. It is intended to investigate this aspect as future work. Another future work is the inclusion of data from other sports present in QASports, such as soccer and football.

\bibliographystyle{plain}
\bibliography{reference}

\end{document}